\pdfoutput=1

\documentclass[11pt]{article}

\usepackage{acl}

\usepackage{times}
\usepackage{latexsym}

\usepackage[T1]{fontenc}

\usepackage[utf8]{inputenc}

\usepackage{microtype}

\usepackage{inconsolata}

\usepackage{booktabs}
\usepackage{linguex}
\usepackage{float}
\usepackage{pgfplots}
\usepackage{color}
\usepackage{multirow}
\usepackage{enumitem}

\pgfplotsset{compat=newest}
\setlist{nosep}

\title{How people talk about each other:\\ Modeling Generalized Intergroup Bias and Emotion}

\author{Venkata S.\ Govindarajan$^1$\ \ \
Katherine Atwell$^2$\ \ \
Barea Sinno$^{3}$
\\
\textbf{Malihe Alikhani}$^2$\ \ \
\textbf{David I.\ Beaver}$^1$\ \ \
\textbf{Junyi Jessy Li}$^1$\\
$^1$ Department of Linguistics, The University of Texas at Austin\\
$^2$ Department of Computer Sciences, University of Pittsburgh\\
$^3$ Department of Political Science, Rutgers University\\
{\small \tt venkatasg@utexas.edu,}
{\small \tt kaa139@pitt.edu,}
{\small \tt barea.sinno@gmail.com,}
\\
{\small \tt malihe@pitt.edu,}
{\small \tt dib@utexas.edu,}
{\small \tt jessy@utexas.edu}\\
}

\begin{document}
\maketitle

\begin{abstract}
Current studies of bias in NLP rely mainly on identifying (unwanted or negative) bias towards a specific demographic group. While this has led to progress recognizing and mitigating negative bias, and having a clear notion of the targeted group is necessary, it is not always practical. In this work we extrapolate to a broader notion of bias, rooted in social science and psychology literature. We move towards predicting interpersonal group relationship (IGR) --- modeling the \emph{relationship between the speaker and the target} in an utterance---using fine-grained interpersonal emotions as an anchor. We build and release a dataset of English tweets by US Congress members annotated for interpersonal emotion -- the first of its kind, and `found supervision' for IGR labels; our analyses show that subtle emotional signals are indicative of different biases. While humans can perform better than chance at identifying IGR given an utterance, we show that neural models perform much better; furthermore, a shared encoding between IGR and interpersonal perceived emotion enabled performance gains in both tasks.
\end{abstract}

\setlength{\Exlabelsep}{0em}
\setlength{\Extopsep}{.2\baselineskip}
\setlength{\SubExleftmargin}{1.3em}

\section{Introduction}
\label{sec:intro}
Currently, most work studying bias in NLP situates bias as negative or pejorative language use towards an individual or group based on traits like race, gender, etc~\citep{kaneko-bollegala-2019-gender, sheng-etal-2019-woman, sap-etal-2020-social, webson-etal-2020-undocumented, Pryzant_DiehlMartinez_Dass_Kurohashi_Jurafsky_Yang_2020, sheng-etal-2020-towards}.  While these approaches greatly advance our understanding of bias in language and its impact and mitigation in NLP, focusing on specific demographic dimensions or an individual's intent is limiting and not always practical. Research in psychology and social science suggests a different perspective. Bias can be seen as a relationship between people and groups, situated in context~\citep{van2009society}; as such, bias refers to differences in behavior (in this case language use) as a result of differences in the relationship between speaker and target. The language we produce is biased in one way or another, whether we intend to or not, and whether that bias is positive, negative, or not clearly associated with any valuation~\citep{beaver2018toward}.

In psychological work on Linguistic Intergroup Bias ~\citep{maass_linguistic_1999}, bias originates from the relationship between the speaker and target of an utterance, i.e. their \textbf{interpersonal dynamics}, and manifests later in  subtle ways. Consider the utterances (tweets) in \ref{ex:inout}, drawn from our collected data in which the identity of the speaker and target are masked:

\ex.\small
\label{ex:inout} \a.\label{ex:ingroup} \textbf{In-group}: We stand w\/ @Doe, who has seen a lot worse than cheap insults from an insecure bully. \#MLKDAY weekend.
\b.\label{ex:outgroup} \textbf{Out-group}: Parents and families live in constant fear for their children with food allergies. A worthy bipartisan cause - thank you @Doe for your leadership on this issue.

\noindent Both express support and admiration towards the target referent \emph{Doe} -- however, the second example uses words indicative that the speaker and target do not share a relevant social identity (in this case, their political party), expressed by words like \emph{bipartisan}. The intensity of admiration expressed is also greater in \ref{ex:ingroup} than \ref{ex:outgroup}. Thus, these two seemingly similar statements differ along interpersonal dimensions that are instructive as to how the bias of the speaker seeps into the utterance.

We now introduce two new tasks that directly model language use in terms of two interpersonal dimensions: (i) \textbf{interpersonal group relationship (IGR) prediction}, where we seek to understand how people talk about others who they consider to be in their same social group (in-group), versus those they consider outside their social group (out-group), and (ii) \emph{perceived} \textbf{interpersonal emotion detection}, where we situate these differences in terms of the emotion expressed in text \emph{towards or in connection with} a target individual described in the utterance. Note that \emph{interpersonal} emotion is different from a more standard, utterance level emotion detection task, as illustrated in row 2 of Table~\ref{tab:emoexs} which has seemingly opposing emotions.

We present a first-of-its-kind, \emph{annotated} dataset for fine-grained interpersonal emotion detection, consisting of 3,033 tweets from members of the US Congress; all of these tweets mention another Congress member, hence providing us with `found supervision' for IGR prediction (whether the speaker and the target belong to the same political party). Our analyses show that while positive interpersonal emotions appear in both in- and out-group situations, negative emotions like anger and disgust are overwhelmingly present in the latter. Meanwhile, human judgments for in vs. out-group membership on this dataset are overly reliant on the polarity of emotion; specifically, human judges are much \emph{less} likely to attribute positive emotions towards out-group targets.

Baseline performances for perceived interpersonal emotion detection shows that this is a challenging task, as is consistent with existing work in emotion detection in general~\cite{demszky-etal-2020-goemotions}. In particular, emotions in this dataset are often expressed with considerable subtlety, likely a characteristic of official political speech. To investigate whether IGR and emotions are intertwined and useful towards each other, we further developed a multi-task model for the prediction of both. We found compelling evidence that multi-tasking IGR and interpersonal emotion improves performance on both tasks with over 10\% improvement in detection of disgust in out-group contexts, and 3\% improvement in IGR prediction.

To summarize the contributions of this paper, we tackle \emph{generalized intergroup bias}, a notion of bias rooted in social psychology that applies to all the various differences in the ways that people talk about others in their in-group or out-group. Standard bias tasks in NLP, and the broader goal of debiasing models could thus be set in a more general context. We present the first dataset to study both interpersonal group membership and emotion, which allows us to analyze both human and model behavior in terms of how the two interact with each other. We release our code and data online at \href{https://github.com/venkatasg/interpersonal-bias}{\texttt{github.com/venkatasg/interpersonal-bias}}.

\begin{table*}[t]
    \small
	\centering
	\begin{tabular}{lll}
		\toprule
		\textbf{Tweet} & \textbf{Interpersonal Emotion} & \textbf{In/Out group?} \\ \midrule
		\multirow{2}{*}{\parbox{9cm}{As @Doe says, the times have found each and every one of us to Defend our Democracy For The People. Worth reading every line.}} & Admiration & In-group \\
		& & \\\midrule
		\multirow{2}{*}{\parbox{9cm}{Freedom has no greater nor tougher champion than @Doe. My prayers are with him and his family.}} & Admiration \& Sadness & In-group \\
		& & \\\midrule
        \multirow{2}{*}{\parbox{9cm}{You don’t get to decide what’s ``fine,'' @Doe. The constitution does. \#DefendOurDemocracy \#WednesdayThoughts}} & Anger \& Disgust & Out-group\\
        &  &  \\\midrule
        \multirow{2}{*}{\parbox{9cm}{Thank you again Senator @Doe for leading the SRF WIN Act[\textellipsis] I'm proud to be a co-sponsor}} & Admiration \& Joy & Out-group  \\
        &  &  \\ \bottomrule
	\end{tabular}
	\caption{Example utterances from our dataset with in/out group and interpersonal emotion labels}
	\label{tab:emoexs}
    \vspace{-4mm}
\end{table*}

\section{Interpersonal Contexts \& Emotions}
\label{sec:setup}
Our aim is to build a generalized, data-driven approach towards studying bias situated in \textbf{interpersonal utterances}, which we define as any utterance where there is a target individual being talked about or referred to. Our goal is to model two novel tasks described below; examples are shown in Table~\ref{tab:emoexs}.

\paragraph{Interpersonal Group relationship} IGR is defined by the relationship between the speaker and target of an utterance. People belong to multiple social groups as part of their identity, however usually only some identities are salient in an utterance in context. We define \emph{in-group} utterances as ones where the speaker and target are in the same social group, and \emph{out-group} utterances as one where they are in different social groups. Given an utterance $u$ written by an individual $s$ with target $t$, the IGR prediction task classifies whether $s$ and $t$ belong to the same social group within the context of $u$.

\paragraph{Interpersonal Emotion} We define \emph{perceived} interpersonal emotion as the emotion expressed by a speaker $s$ \emph{towards, or in connection with} the target $t$ of the utterance $u$, as perceived by a reader.  We use the Plutchik wheel of emotions, which is widely adopted in the community, as the basis of our emotion taxionomy~\cite{plutchik2001nature}; we use the 8 fundamental  emotions (\emph{admiration, anger, disgust, fear, interest, joy, sadness, surprise}) instead of the full 24 emotions in the wheel due to data sparsity. Interpersonal emotions may be different, or a subset of, emotion for the whole of an utterance, as illustrated in rows 2, 3 and 4 of Table~\ref{tab:emoexs}. Given an utterance $u$ written by an individual $s$ with target $t$, the interpersonal emotion detection task identifies the perceived emotion of $s$ towards the target $t$.

\section{Data Collection}
\label{sec:data}
In our area of focus, we require natural language data which satisfies the following criteria: \emph{(1)} Each utterance must have at least one target about whom the utterance mainly concerns. \emph{(2)} The relationship between the speaker and the target must be inferred based on metadata or other information. Specifically, we are interested in aspects of their social identity that they share or differ on.

The dataset we collect comes from tweets by members of US Congress where other members are mentioned in the same tweet. We use this as a convenient testbed: each member's group affiliation (i.e., their party identity) is public, thus we can easily know whether the speaker is tweeting to a target in their own party or not.\footnote{For simplicity, we do not consider other factors such as the home state of a congress member.} In other words, this dataset gives us ``found supervision'' for our first task of IGR prediction. For our second task, we annotate a subset of these tweets for perceived interpersonal emotion; this is, to our knowledge, the first dataset dedicated to interpersonal emotion.

\subsection{Data Sources and Preprocessing}
\label{subsec:datasources}

Social media text like tweets offer a fertile ground for our study. A focus on tweets with \emph{mentions} in them satisfies our first criterion -- people generally use mentions to say something about or towards another individual on twitter. Tweets by members of US Congress are a matter of public record, and we can infer the social relationship (in terms of party affiliation) between speaker and target using publicly available information. We prioritize working with a dataset of tweets by members of US Congress (downloaded using the Twitter API) between 2010 and 2021, spanning two presidencies, during which both parties held power in Congress. We filter these tweets to exclude retweets, and include those tweets that mention \emph{at most} one other member of Congress whose party affiliation is known. We believe these 2 assumptions are sufficient to arrive at a dataset of tweets where the speaker is talking towards/about \emph{one} target. Thus, we restrict ourselves to two social groups in this sphere --- Democrat and Republican parties in the US. We sample an equal number of in-group and out-group tweets from a large sample consisting of all tweets by members of Congress. Apart from years 2010--2012 and 2021 which contained fewer tweets due to sparsity issues, we sampled at least 300 tweets each year.

\subsection{Interpersonal Emotion Annotation}

While we can infer whether a tweet is in-group or out-group based on the identity of speaker and target whose political affiliations are known, we still require annotated data on perceived interpersonal emotions. Interpersonal emotions vary in subtle ways from sentiment or overall sentiment of utterances: an utterance can have negative sentiment overall, but still convey positive emotions towards the target of the sentence (expressing admiration at someone's death for instance). For this reason, we devise an annotation schema for annotating \emph{the emotion expressed by speaker $s$ towards target $t$}.

\paragraph{Instructions} Annotators are presented with a tweet, with the identity of the speaker unknown and that of the target masked with a placeholder name \textbf{@Doe} to minimize potential biases of the annotators' prior knowledge of party affiliation intruding into the annotation:

\ex.\label{ex:doe} If \textbf{@Doe} can get her hair done in person, Congress can vote in person\textellipsis

Annotators are instructed to read the tweet and select only the most notable emotion(s) they think are expressed by the tweet author \emph{in connection with} \textbf{@Doe}. To aid annotators, we provide examples of the 8 Plutchik emotions (\emph{joy, admiration, fear, suprise, sadness, digust, anger and interest}) expressed as interpersonal emotions in tweets. Annotators are also shown a schematic of the Plutchik wheel of emotions, which acquaints them with how the emotions are related to one another in our framework. Annotators are allowed to select more than one emotion to account for emotion co-occurrence. We also explicitly tell annotators that more than one of the emotions can be present in the tweets, to encourage them to select all interpersonal emotions expressed. They are also allowed to not choose any emotion.

\paragraph{Annotation}  To obtain reliable annotations, we prequalify annotators using a qualifying task. Annotators were recruited on Mechanical Turk using a qualifying task where they were asked to annotate 6 tweets using the schema shown above. We restricted the qualification task to annotators living in the USA who had attempted at least 500 HITS and had a HIT approval rate $\geq$ 98\%. After manual inspection, 6 annotators were qualified for bulk annotation. Each tweet was annotated by three different annotators. To ensure annotators were paid a fair wage of at least 10\$ an hour, we paid annotators \$0.50 per HIT. Each HIT involved annotating 3 tweets, which we estimate to take on average 3 minutes to complete.  In total, 3,033 tweets between 2010 and 2021 were annotated with perceived interpersonal emotion.

\begin{figure}[t]
	\centering
	\includegraphics[width=0.7\linewidth]{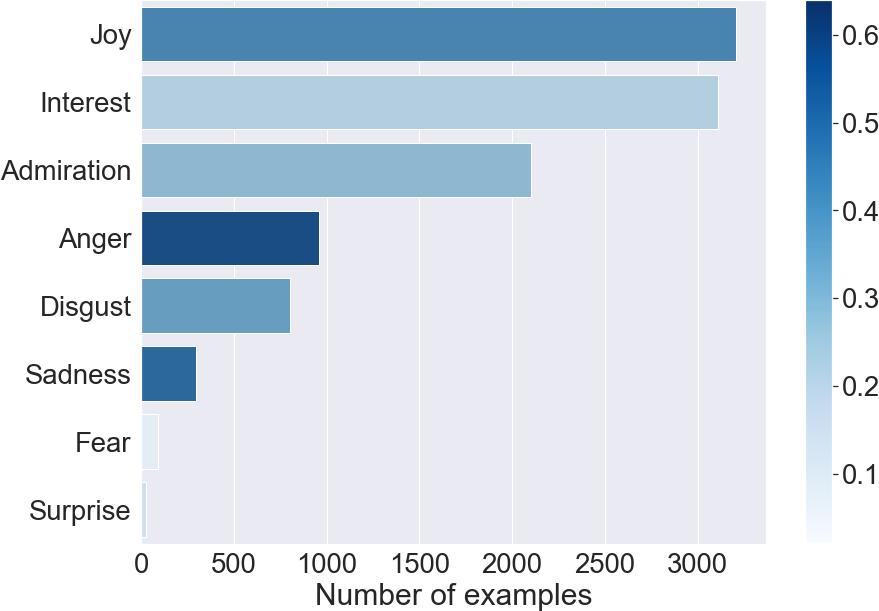}
	\caption{Emotions ordered by the number of examples where at least one rater uses a particular label. The color indicates the average interrater correlation.}
	\label{fig:interrater-corr}
    \vspace{-4mm}
\end{figure}

\paragraph{Agreement}   To measure agreement between annotators on the Plutchik-8 emotion wheel, we use the Plutchik Emotion Agreement (PEA) score from \citet{desai-etal-2020-detecting}. The PEA score addresses the issue of penalizing all disagreements equally, by penalizing dissimilar emotion annotations higher than more similar ones (according to the Plutchik wheel). Our PEA score is 0.73. The original PEA formulation used the best(max) pair of emotion annotations between two workers. Taking the \emph{worst} combination of emotions between two workers (averaged over all tweets and workers), the PEA (min) score is 0.60. Overall, we find moderate to high agreement on fine-grained interpersonal emotions.  In Figure~\ref{fig:interrater-corr} we also present interrater correlation, a metric used in~\citet{demszky-etal-2020-goemotions}; we see that distributions are similar.

\begin{table}[t]
	\centering
	\small
	\begin{tabular}{llll}
		\toprule
		\textbf{Emotion} & \textbf{Train} & \textbf{Dev} & \textbf{Test} \\ \midrule
		Admiration & 467 & 64 & 58\\ 
        Anger & 225 & 40 & 46\\ 
        Disgust & 206 & 32 & 43\\ 
         Fear & 1 & 0 & 0\\ 
        Interest & 701 & 83 & 84\\ 
        Joy & 801 & 107 & 106\\ 
        Sadness & 72 & 11 & 11\\ 
         Surprise & 1 & 0 & 0\\ 
         \emph{No Emotion} & 519 & 56 & 63\\\bottomrule
	\end{tabular}
	\caption{Distribution of emotions in train-dev-test split}
	\label{tab:traindevtestsplit}
\end{table}

\paragraph{Aggregation} We consider a tweet to have a certain emotion label if at least 2 out of 3 annotators agree that the particular emotion was present in the tweet. A total of 638 tweets have no interpersonal emotion associated with them. We employ a 80-10-10 train-dev-test split on our data.

The number of annotated examples (tweets) per emotion is shown in Table~\ref{tab:traindevtestsplit}. We omit \emph{fear} and \emph{surprise} from future tables due to the absence of annotated examples.

\section{Preliminary Analysis}
\label{sec:prelimanalysis}
\paragraph{How are emotions distributed?} When observing the distribution of aggregated emotion labels themselves, a clear pattern emerges as seen in Table~\ref{tab:inoutdist}. Negative emotions such as anger and disgust are almost always expressed in out-group settings, while positive emotions are present in both in-group and out-group settings. A similar distribution of emotions was observed for Democrats and Republicans --- members of both parties reserved their public anger and disgust for members of the other party. This reflects an innate bias in terms of the distribution of interpersonal emotions per situation, and warrants future work to explore negative interpersonal emotions in an in-group setting.

\begin{table}[t]
	\centering
    \small
	\begin{tabular}{llll}
		\toprule
		\textbf{Emotion} & \textbf{All} & \textbf{In-Group} & \textbf{Out-Group} \\ \midrule
		Admiration & 15.5 & 22.2 & 9.1\\ 
        Anger & 8.2 & 1.0 & 15.1\\ 
        Disgust & 7.4 & 0.3 & 14.2\\ 
        Interest & 22.9 & 27.2 & 18.6\\ 
        Joy & 26.7 & 32.2 & 21.4\\ 
        Sadness & 2.5 & 2.6 & 2.4\\ 
        \emph{No Emotion} & 16.8 & 14.5 & 19.1 \\
        \bottomrule
	\end{tabular}
	\caption{Proportion of emotions in different interpersonal contexts}
	\label{tab:inoutdist}
\end{table}

\begin{figure}[t]
	\centering
	\includegraphics[width=0.7\linewidth]{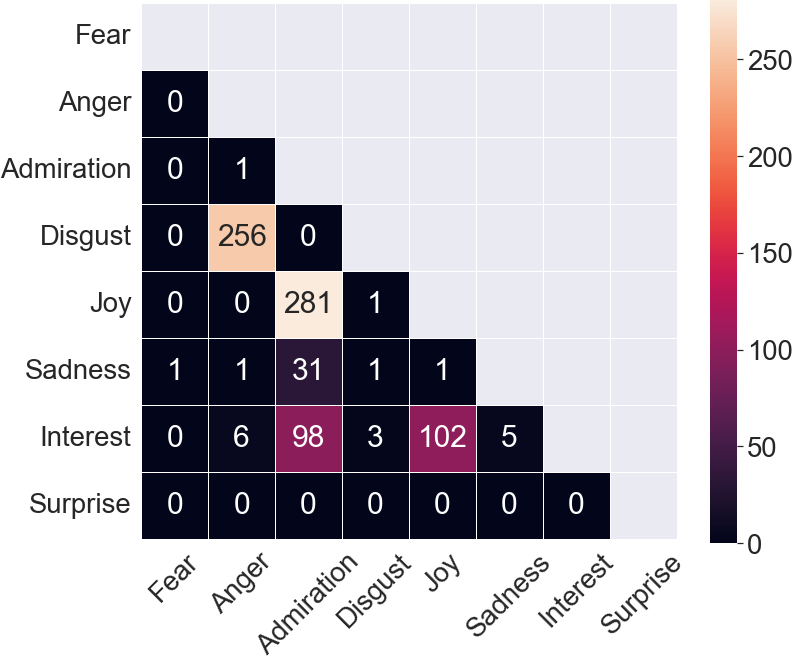}
	\caption{Co-occurence of emotions in our dataset.}
	\label{fig:heatmap}
\end{figure}

Figure~\ref{fig:heatmap} shows the co-occurrence of interpersonal emotions in our dataset. We can see that emotions that are farther apart and more dissimilar, such as admiration and disgust, joy and sadness, co-occur infrequently. Emotions that are closer such as anger and disgust, admiration and joy, co-occur much more often. The only outlier is the higher than normal co-occurrence of admiration with sadness --- after a closer examination, this can be attributed to tweets expressing admiration and sadness at the passing, or end of the career, of a fellow congressperson.

\paragraph{Who were the targets of negative emotions?}

On further analysis, it appears that most of the out-group disgust and anger is directed at 3 handles -- @speakerryan, @speakerpelosi, and @speakerboehner who were all Speakers of the House of Representatives over most of the time period of our dataset. 63.7\% of disgust and 64.3\% of anger is directed towards these three twitter handles. 11.9\% of all tweets in our dataset are directed at these handles, indicating the preponderance of negative interpersonal emotion directed at the Speaker of the house. However, we note that negative emotions like anger and disgust were still expressed towards 51 and 45 different individuals in our dataset, respectively.

\begin{figure*}[t]
	\centering
	\includegraphics[width=\linewidth]{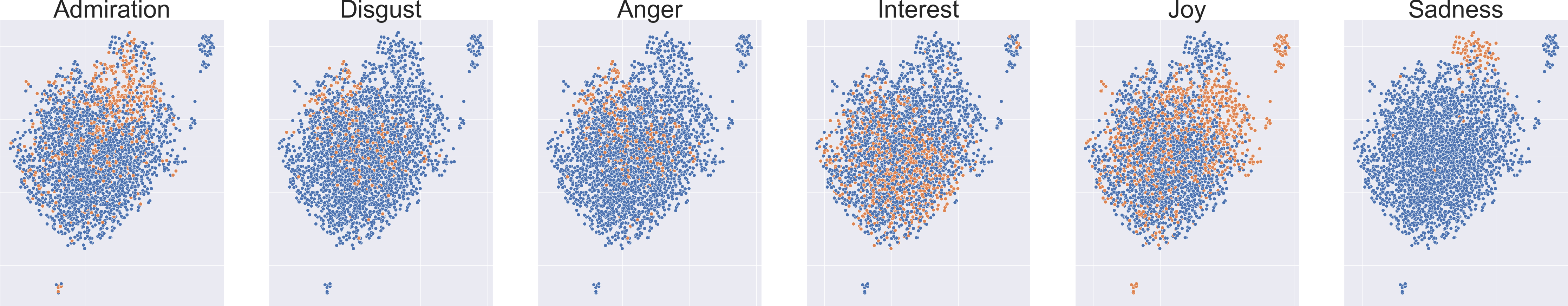}
	\caption{Distribution of interpersonal emotions in unsupervised representations of tweets in our dataset. Orange indicates the emotion was present for that tweet. Each point represents one tweet from our dataset.}
	\label{fig:domainclus}
    \vspace{-2mm}
\end{figure*}

\paragraph{Can humans predict in/out-group?}

While our data naturally comes with ``gold'' IGR labels, what is unexplored is whether the distinction between in-group and out-group speech is prominent and noticeable by humans. Additionally, it is also unclear if humans might have their own expectation of how in/out-group speech should be characterized.

Concretely, we investigate if human annotators were capable of accurately performing the IGR prediction task when the speaker and target are masked. Two authors of this paper, one a social science graduate student, and the other a computational linguistics graduate student, annotated 50 random tweets from our validation data which they had not been exposed to earlier for in/out group labels. Their Fleiss $\kappa$ agreement score was 0.64, indicating moderate agreement.

To check how accurate their judgements were, we calculate for each annotator their F1 score against our ``gold'' in/out group labels. Their F1 scores on these 50 tweets were 0.67 and 0.63, which as we will discuss in Section~\ref{sec:results}, only match simple baselines of supervised systems. Annotators comments indicate that they overly relied on the sentiment of tweets to make the classification --- positive sentiment means in-group and negative sentiment means out-group. While negative emotions are over-represented in out-group situations as Table~\ref{tab:inoutdist} shows, our dataset contains a substantial presence of out-group tweets with positive interpersonal emotions as well. Annotators also noticed some lexical cues like `bipartisan' that are indicative of out-group tweets.

\paragraph{Do pre-trained representations capture interpersonal emotions?}

Pre-trained language models have been found to learn sentence representations that cluster by domain without supervision~\citep{aharoni-goldberg-2020-unsupervised}. We wished to investigate if any of our annotated properties cluster inherently in reduced representations of the tweets in our data. To obtain unsupervised representations, we use BERTweet~\citep{nguyen_bertweet_2020}, a language model pre-trained on 850M English tweets. We take the 768 dimensional embeddings from the final layer of the \texttt{<s>} token in BERTweet, and dimensionally reduce them to 2 dimensions using UMAP~\citep{sainburg2021parametric}. Figure~\ref{fig:domainclus} shows the distribution of tweets, color coded for interpersonal emotions. While there is a lot of overlap between representations when stratified by emotion, we can see that some emotions that are intuitively opposite, like admiration \& disgust, joy \& sadness are moderately separable. This indicates that interpersonal emotions do define some topic or domain level properties of a tweet.

\section{Experiments}
\label{sec:models}
We detail our experiments for the two novel tasks discussed in Section~\ref{sec:setup}: predicting the IGR (in-group or out-group) given a tweet, and predicting the interpersonal emotion given a tweet. We present baselines for the two tasks separately, and also present a multi-task model to gauge the extent to which knowledge of IGR may help in predicting interpersonal emotion, and vice versa.

\subsection{Interpersonal Group Relationship}

\paragraph{Sentiment-Rule} Our first baseline is a rule-based one leveraging coarse sentiment: if a tweet's sentiment is predicted to be negative, classify it as out-group; if positive, classify it as in-group; and if neutral, classify it as either in-group or out-group randomly. We use a RoBERTa-Base model finetuned for sentiment on tweets~\citep{barbieri-etal-2020-tweeteval} to extract the sentiment of each tweet in our dataset.
	
\paragraph{NB-SVM} As a second baseline, we build an SVM model that uses Naive-Bayes log-counts ratios of unigrams and bigrams~\citep{wang-manning-2012-baselines}.

\paragraph{BERTweet} We use BERTweet~\citep{nguyen_bertweet_2020}, a language model pre-trained on 850M English tweets as our dataset consists purely of English language tweets. A classification head is placed on top of the language model. We also experiment with a version where the language model parameters are frozen, and only the classification head parameters are finetuned (BERTweet-ft).

The input to all models is only the tweet with no other context, and the target masked with a placeholder \texttt{@USER}.

\subsection{Interpersonal Emotion}

\paragraph{EmoLex} As a baseline model for interpersonal emotion identification, we rely on EmoLex~\citep{Mohammad2013CROWDSOURCINGAW}. EmoLex consists of 14,182 crowdsourced words associated with the 8 basic Plutchik emotions. Critically, these words appear in emotional contexts, but are not necessarily emotion words themselves. EmoLex counts occurrences of words from its lexicon in an utterance, and assigns a normalized score for each emotion based on occurrence frequency. We consider an emotion to be on, if it's normalized score is $\geq$ 0.001. While EmoLex has issues with regards to its context insensitivity and the social biases built into its lexicon~\citep{zad-etal-2021-hell}, we include it as a baseline to understand to what extent interpersonal emotions can be deduced using a lexicon.

\paragraph{BERTweet} We use the same BERTweet model as earlier. We add a dense output layer on top of the pretrained model for the purposes of finetuning, with a sigmoid cross entropy loss function to support multi-label classification. The loss is weighted for each of the 8 emotion labels with the ratio of positive and negative examples to increase precision. If none of the 8 emotion labels are flipped on, we consider that to be the `No Emotion' label, i.e. there is no interpersonal emotion between speaker and target in the tweet. We experiment with a version of the model where the language model parameters are frozen and only the labelling head parameters are finetuned (BERTweet-ft).

\subsection{Multi-Task Model}

In \textsection~\ref{sec:prelimanalysis}, we observed that the emotions anger and disgust are overwhelmingly present in out-group situations. Thus, we hypothesize that IGR information would be useful towards interpersonal emotion identification, and vice versa. To test this hypothesis we train a multi-task model. The model is trained to predict \emph{both} the IGR label and emotion using shared parameter finetuning.

We use the same BERTweet model as earlier. We add two dense output layers on top of the pretrained model, one for classifying IGR and another for labelling interpersonal emotion. Both heads share the same parameters below. These are trained with same loss as earlier individual models. The model alternates between finetuning for group relationship and emotion over every training item.

\subsection{Implementation}
We use \texttt{bertweet-base} pretrained embeddings from Huggingface's models hub~\citep{wolf-etal-2020-transformers}.
All models are finetuned for a maximum of 20 epochs with early stopping. Early-stopping patience for models trained on each task separately is 3. The patience for the multi-task model is set at 5 as the multi-tasking setup led to slower convergence. The learning rate for the classification heads was set at 5e-3 while the learning rate for the internal language model parameters was set at 2e-5. Dropout probabilities in classification heads was set at 0.1. The best performing model before early stopping on validation data was chosen in all cases. We report F1 scores averaged over 3 random restarts for all models, with the standard deviation in parentheses next to the mean.

\section{Results and Analysis}
\label{sec:results}
\begin{table}[t]
    \small
	\centering
	\begin{tabular}{ll|ll}
		\toprule
		\textbf{Model} & \textbf{F1} & \textbf{Model} & \textbf{F1} \\ \midrule
		Majority class & 51.1 & BERTweet & 74.1 (3.3) \\
		Sentiment-Rule & 56.3 & BERTweet-ft & 66.5 (1.6) \\
		NB-SVM & 62.5 & Multitask & 77.3 (0.8) \\
        Human & 66.7 & & \\
        \bottomrule
	\end{tabular}
	\caption{Results on test set, with SD in parentheses, for interpersonal group relationship prediction task.}
	\label{tab:inoutresults}
\end{table}

\begin{table}[t]
    \centering
    \small
    \begin{tabular}{ll}
        \toprule
        \textbf{In-group} & \textbf{Out-group} \\ \midrule
        thanks, love, count me & thanks, bipartisan, restore \\
        birthday, my colleague & kind, resignation \\\bottomrule
    \end{tabular}
    \caption{Top unigram and bigram features from NB-SVM model for each class.}
    \label{tab:svm-features}
\end{table}

\begin{table}[t]
    \footnotesize
    \setlength{\tabcolsep}{2.5pt}
	\centering
	\begin{tabular}{lllll}
		\toprule
		\textbf{} & \textbf{Emo} & \textbf{BERTweet} & \textbf{BERTweet} & \textbf{Multi-} \\
		& \textbf{Lex} & & \textbf{-ft} & \textbf{task} \\ \midrule
		Admir. & 37.5 & 70.3 (3.7) & 40.7 (1.1) & 68.9 (1.6) \\
		Anger & 26.6 & 71.3 (11.2) & 23.0 (3.4) & 69.3 (3.3) \\
		Disgust & 25.5 & 47.1 (21.6) & 13.0 (4.1) & 74.5 (7.1) \\
		Interest & 0 & 53.1 (3.3) & 5.8 (2.4) & 51.5 (8.5) \\
		Joy & 48.4 & 85.9 (1.9) & 71.3 (1.4) & 83.6 (1.3) \\
		Sadness & 4.3 & 11.1 (9.6) & 0 & 33.6 (18.5)\\
		\emph{No Emotion} & 22.2 & 49.1 (1.2) & 43.4 (3.8) & 71.6(1.2) \\\bottomrule
	\end{tabular}
	\caption{F1 scores on test set, with SD in parentheses, for interpersonal emotion labelling task.}
	\label{tab:emoresults}
\end{table}

\paragraph{Interpersonal Group relationship} In modeling IGR, we find that \texttt{Sentiment-Rule} performs not much better than chance (Table~\ref{tab:inoutresults}). This underscores one strength of our data, which contains a sizable number of out-group tweets with positive interpersonal emotion attached to them. The \texttt{NB-SVM} model based on unigrams and bigrams performs slightly better, and picks up on some obvious out-group lexical cues like the lemma `bipartisan', as shown in Table~\ref{tab:svm-features}. The \texttt{BERTweet} model performs substantially better, performing over 10 points better than humans. The model, with only the classification head finetuned, leaving the language model parameters intact(\texttt{BERTweet-ft}) performs about 10 points above chance --- indicating that there may be features advantageous towards this task in the vanilla LM itself.

\paragraph{Interpersonal Emotion}

We find that the \texttt{EmoLex} baseline, which relies purely on lexical cues, performs dismally on our data, with poor performance in both in-group and out-group settings(Table~\ref{tab:emoresults}). This is a strong indication that emotions are expressed more implicitly in this dataset. The \texttt{BERTweet} model performs substantially better, indicating that interpersonal emotions, even if implicit, can be learned.

\begin{table}[t]
	\centering
	\small
	\begin{tabular}{lll}
		\toprule
		\textbf{Emotion} & \textbf{BERTweet} & \textbf{MultiTask} \\\midrule
		Admiration & 77.9 (2.6) & 72.8 (3.9)\\
        Anger & 71.7 (9.9) & 69.4 (3.4)\\
        Disgust & 48.2 (22.4) & 75.9 (6.5)*\\\bottomrule
	\end{tabular}
	\caption{F1 scores on test set, SD in parentheses on out-group tweets. * indicates statistical significance (p<0.05)}
	\label{tab:detemoresults}
\end{table}

\paragraph{Multitask Model}

As Table~\ref{tab:inoutresults} shows, Multi-tasking the two tasks leads to a noticeable improvement in F1 for IGR prediction, with the differences being statistically significant using a bootstrap test~\cite[p$<$0.05;][]{berg-kirkpatrick-etal-2012-empirical}; the multi-task model is also more stable with much lower variance across runs. These results suggest that interpersonal emotion is useful towards IGR prediction.

Table~\ref{tab:emoresults} shows that the performance of the multitask model on predicting interpersonal emotions is significantly better that the BERTweet model (p$<$0.05) on emotions like \emph{disgust}, which suggests that IGR is useful towards the task of emotion identification. Furthermore, multitasking boosted performance at predicting the \emph{no emotion} label by 20\%.  Table~\ref{tab:detemoresults} compares the multitask model's performance against the BERTweet model in \emph{out-group} settings (where most of the gains were found) for 3 emotions --- illustrating the boost in performance afforded by joint modeling of IGR and emotion for \emph{digust}. The 3 emotions listed also showed significant differences in their distribution in in-group and out-group settings.

\paragraph{Humans vs.\ Models}

Comparatively, we find that model performance exceeds human performance on the task of in-group versus out-group prediction, albeit not on the same dataset. The model's main driver of performance is its high accuracy on positive intergroup emotion out-group tweets, such as those expressing admiration or joy. Human annotators consistently fall back on the heuristic that sentences with positive affect probably imply that the speaker is talking about someone in their in-group. But it is not the case in the political domain, where overtures to bipartisanship serve as useful signals. For instance, both \ref{ex:adm-1} and \ref{ex:adm-2} express admiration towards the target Doe, where the first is in-group while the second is out-group. The call to civility is the only subtle linguistic cue that this tweet may constitute out-group speech.

\ex. \a.\label{ex:adm-1} Admire @OfficialCBC Chairman @Doe's moral voice on issues of racism and restorative justice. He is a real leader for our nation and Congress.
     \b.\label{ex:adm-2} A decade has passed, but our friendship is the same. Proud to work with @Doe to \#ReviveCivility. \#tbt Read more about our efforts here:
     
Future work needs to look into what information the embeddings are using to make their classification decision.

\paragraph{Model Errors} While the multitasking setup improves model performance on the task of predicting IGR, and outperforms human labelers in our small pilot, it still gets some easy examples wrong, such as labelling~\ref{ex:disgust-ex} as in-group even though it expresses some disgust at the target. The model also falls into the same trap as human labelers --- for instance assuming that a tweet expressing admiration must be in-group~\ref{ex:admiration-ex}.

\ex.\label{ex:disgust-ex} Trump selected @USER for HHS Secretary. Price has undeniable history of cutting access to healthcare to millions, especially women.

\ex. \label{ex:admiration-ex} Inspiring speech from @USER - we have a duty to represent our country with respect \& dignity. \#NationalDayofCivility.

To ensure that model performance on IGR prediction is not limited by the size of our training data, we experimented with training \texttt{BERTweet} models on larger datasets. Since we have `found supervision' for IGR labels, we only need to increase training data size by sampling more tweets from relevant accounts using the same procedure detailed in \textsection~\ref{subsec:datasources}. We found that F1 score does not increase with more training data.

Future work needs to look into linguistically motivated ways to improve model performance on the IGR task. Since we have observed that the multi-task setup improves model performance, perhaps other multi-task setups, such as modeling the overall affect towards the target expressed by the speaker might help in modeling IGR better.

\section{Related Work}
\label{sec:background}
\paragraph{Emotion and Stance Detection} A wealth of work has looked at corpora and models for the detection of perceived emotion in social media text~\citep{mohammad-2012-emotional,6406313, Mohammad2015UsingHT, abdul-mageed-ungar-2017-emonet,desai-etal-2020-detecting,demszky-etal-2020-goemotions}. However existing work doesn't distinguish between emotion of a sentence as a whole, versus interpersonal emotion towards a target. The task closest to our study of interpersonal emotions is stance detection: whether the author has a favourable, neutral, or negative position towards a proposition or target. \citet{mohammad-etal-2016-semeval} looked at stance in five target domains are given: abortion, atheism, climate change, feminism and Hillary Clinton. While stance detection focuses on a collection of utterances with the same topic, our interest is in modeling interpersonal emotion towards a target individual which is more fine-grained and can vary in each utterance.

\paragraph{Intergroup bias in Psychology} The  Linguistic Intergroup Bias  (LIB) theory~\cite{maass_language_1989,maass_linguistic_1999} states that there is a systematic asymmetry in language production qualities of a speaker as a function of the social category to which the referent of an utterance belongs. Through psycholinguistic experiments, LIB seeks to explain why stereotypes are transmitted and persist in daily life: in an interpersonal situation, socially desirable in-group behaviors and undesirable out-group behaviors are encoded at a higher level of \textbf{abstraction}, whereas socially undesirable in-group behaviors and desirable in-group behaviors are encoded at a lower level of abstraction. Work in psychology and psycholinguistics reproduced LIB in various domains such as political news reporting~\cite{Anolli2006LinguisticIB} and crime reporting~\cite{gorham_news_2006}; as well as work exploring how LIB can be used as an indicator for a speaker’s prejudicial attitudes~\cite{hippel_linguistic_1997}, or as a predictor for racism~\cite{schnake_modern_1998}. 

Contemporaneous studies on LIB, however, are hand-coded and have so far tended to focus on narrow concepts such as abstractness of the verb and coarse notions of sentiment. Nonetheless, the LIB hypothesis connects the two dimensions of interpersonal dynamics studied here with a third dimension directly related to semantic properties of the utterance.

\section{Conclusion}
\label{sec:conclusion}
Taking a cue from studies of bias in social science and psychology, we situate bias in language use through the lens of interpersonal relationships between the speaker and target of an utterance, and the speaker's interpersonal emotional state with respect to the target. Over a corpus of tweets by members of US Congress,  we introduce two novel tasks -- interpersonal group relationship prediction (IGR) and interpersonal emotion labelling, to better understand variation in language as a function of social relationship between speaker and target in interpersonal utterances. We find certain interpersonal emotions like anger and disgust are over-represented in out-group situations, with the majority of the negative emotions directed at leaders of the two political parties. Through modeling studies, we find that transformer based models perform better than humans at predicting IGR given an utterance, raising the question as to what latent features of language the model uses to make this decision. Finally, we also find that joint modelling of the two dimensions is beneficial to prediction of certain interpersonal emotions in out-group situations. Future work needs to look into what information is useful for predicting IGR and emotions -- with the Linguistic Intergroup Bias literature offering a clue as to which higher level semantic features vary systematically.

\section*{Ethics Statement}
\label{sec:ethics}
For our corpus of tweets on which we performed annotations, we downloaded the tweets using the official Twitter API. In accordance with the Twitter Terms of Service, we release tweet IDs and usernames, but not the tweet text itself. Our dataset was built through crowdsourced annotations on Amazon Mechanical Turk. To ensure annotators were paid a fair wage of at least \$10 an hour, we paid annotators \$0.50 per HIT. Each HIT involved annotating 3 tweets, which we estimate to take on average 3 minutes to complete.

\section*{Limitations}
\label{sec:limitations}
Our results show the importance of having reliable and accurate emotion prediction models, which is an open problem in psychology and computer science. Future work might look into identifying different fine-grained emotional constructs and study their correlations with the underlying linguistic biases. Future work may also look into the generalizability of the results presented here in other domains of language use.

While we present the utterances as constituting natural speech by one speaker (the congressperson who sent the tweet), it is likely most congresspeople employ social media teams that help in crafting the language of some of their tweets. However, we believe for the sake of interpersonal group membership, the relationship between the speakers and their targets would not be affected.

Finally, while we show that transformer based models perform better at IGR prediction than humans, we note that the human performance was on a small subset of test data. While it is possible that these models discovered latent features that could explain their better performance, the model could also be using spurious features idiosyncratic to our dataset, rather than true differences in in-group versus out-group speech.

\section*{Acknowledgements}
This research is partially supported by NSF grants IIS-2107524, IIS-2145479 and Good Systems,\footnote{ \url{http://goodsystems.utexas.edu}} a UT Austin Grand Challenge to develop responsible AI technologies. We also acknowledge the Texas Advanced Computing Center (TACC)\footnote{\url{https://www.tacc.utexas.edu}} at UT Austin for providing the computational resources for many of the results within this paper.

\bibliography{references}
\bibliographystyle{acl_natbib}

\end{document}